\documentclass[conference]{IEEEtran}
\IEEEoverridecommandlockouts
\usepackage{cite}
\usepackage{hyperref}
\usepackage{amsmath,amssymb,amsfonts}
\usepackage{algorithmic}
\usepackage{graphicx}
\usepackage{textcomp}
\usepackage{xcolor}
\def\BibTeX{{\rm B\kern-.05em{\sc i\kern-.025em b}\kern-.08em
    T\kern-.1667em\lower.7ex\hbox{E}\kern-.125emX}}

\usepackage{booktabs}
\usepackage{multirow}

\ifCLASSOPTIONcompsoc
  \usepackage[caption=false,font=normalsize,labelfont=sf,textfont=sf]{subfig}
\else
  \usepackage[caption=false,font=footnotesize]{subfig}
\fi

\begin{document}

\title{SAM-FNet: SAM-Guided Fusion Network for Laryngo-Pharyngeal Tumor Detection\\
}

\author{\IEEEauthorblockN{1\textsuperscript{st} Jia Wei\IEEEauthorrefmark{1}\thanks{This work is partially supported by the Basic and Applied Basic Research Project of Guangdong Province (2022B1515130009), the Special subject on Agriculture and Social Development, Key Research and Development Plan in Guangzhou (2023B03J0172), and the Natural Science Foundation of Top Talent of SZTU (GDRC202318).}}
\IEEEauthorblockA{\textit{College of Big Data and Internet} \\
\textit{Shenzhen Technology University}\\
Shenzhen, China \\
provvvj@gmail.com}
\and
\IEEEauthorblockN{2\textsuperscript{nd} Yun Li\IEEEauthorrefmark{1}\thanks{Jia Wei and Yun Li are the co-first authors.}}
\IEEEauthorblockA{\textit{Otorhinolaryngology Hospital} \\
\textit{The First Affiliated Hospital, Sun Yat-Sen University}\\
Guangzhou, China \\
liyun76@mail.sysu.edu.cn}
\and
\IEEEauthorblockN{3\textsuperscript{th} Meiyu Qiu}
\IEEEauthorblockA{\textit{School of Applied Technology} \\
\textit{Shenzhen University}\\
Shenzhen, China \\
qiumeiyu2023@email.szu.edu.cn}
\and
\IEEEauthorblockN{4\textsuperscript{rd} Hongyu Chen}
\IEEEauthorblockA{\textit{College of Big Data and Internet} \\
\textit{Shenzhen Technology University}\\
Shenzhen, China \\
202200202155@stumail.sztu.edu.cn}
\and
\IEEEauthorblockN{5\textsuperscript{th} Xiaomao Fan\IEEEauthorrefmark{2}}
\IEEEauthorblockA{\textit{College of Big Data and Internet} \\
\textit{Shenzhen Technology University}\\
Shenzhen, China \\
astrofan2008@gmail.com}
\and
\IEEEauthorblockN{6\textsuperscript{th} Wenbin Lei\IEEEauthorrefmark{2}\thanks{Xiaomao Fan and Wenbin Lei are the corresponding authors.}}
\IEEEauthorblockA{\textit{Otorhinolaryngology Hospital} \\
\textit{The First Affiliated Hospital, Sun Yat-Sen University}\\
Guangzhou, China \\
leiwb@mail.sysu.edu.cn}
}

\maketitle

\begin{abstract}
Laryngo-pharyngeal cancer (LPC) is a highly fatal malignant disease affecting the head and neck region. Previous studies on endoscopic tumor detection, particularly those leveraging dual-branch network architectures, have shown significant advancements in tumor detection. These studies highlight the potential of dual-branch networks in improving diagnostic accuracy by effectively integrating global and local (lesion) feature extraction. However, they are still limited in their capabilities to accurately locate the lesion region and capture the discriminative feature information between the global and local branches. To address these issues, we propose a novel SAM-guided fusion network (SAM-FNet), a dual-branch network for laryngo-pharyngeal tumor detection. By leveraging the powerful object segmentation capabilities of the Segment Anything Model (SAM), we introduce the SAM into the SAM-FNet to accurately segment the lesion region. Furthermore, we propose a GAN-like feature optimization (GFO) module to capture the discriminative features between the global and local branches, enhancing the fusion feature complementarity. Additionally, we collect two LPC datasets from the First Affiliated Hospital (FAHSYSU) and the Sixth Affiliated Hospital (SAHSYSU) of Sun Yat-sen University. The FAHSYSU dataset is used as the internal dataset for training the model, while the SAHSYSU dataset is used as the external dataset for evaluating the model's performance. Extensive experiments on both datasets of FAHSYSU and SAHSYSU demonstrate that the SAM-FNet can achieve competitive results, outperforming the state-of-the-art counterparts. The source code of SAM-FNet is available at the URL of \href{https://github.com/VVJia/SAM-FNet}{https://github.com/VVJia/SAM-FNet}.
\end{abstract}


\section{Introduction}
Laryngo-pharyngeal cancer (LPC) is a malignant disease with a high mortality in head and neck tumor. Reports \cite{sung2021global} on 2020 showed that LPC have caused more than 130,000 deaths globally. Early-stage LPC can often be effectively treated using minimally invasive procedures, boasting a 5-year survival rate of up to 90\% with perserving patient's voice \cite{rudolph2011effects}. Clinically, laryngologists typically rely on biospy under laryngoscope as the gold standard \cite{sampieri2024real} for LPC diagnosis. However, the visual inspection process is time-consuming and subjects to laryngologist's skill and experience\cite{2011irjala, azam2022deep}. Limitations of the laryngologist's expertise may lead to missed diagnoses and unnecessary repeated biopsies \cite{ni2019clinical, chen2023accuracy}. Therefore, developing an automatic approach to assist the laryngologists to detect LPC is of great significance. 

In recent studies, many researches attempted to utilize deep learning techniques to build models for tumor detection. These methods are mainly categorized into two groups: single-branch based networks \cite{U-Net, chen2017deeplab, Efficientnet, ViT, UC-DenseNet, MTANet, medsam} and dual-branch based networks \cite{DLGNet, RadFormer, DSI-Net, DSMT-Net, yang2024cvan}. As for the single-branch based network, the existing methods focus solely on the global features of the endoscopic images as input. For instance, Luo \emph{et al.} \cite{UC-DenseNet} proposed a UC-DenseNet combining Convolution Neural Network (CNN) and Recurrent Neural Network (RNN) to recognize ulcerative colitis, and also utilized attention mechanism to localize the relevant feature information. Ling \emph{et al.} \cite{MTANet} developed a multi-task attention network termed MTANet based on Transformer and CNN architecture, which achieved excellent performance in the detection of multiple types of tumors. Though these methods have achieved promising results, they overlook the valuable local information within lesion regions, which is informative for accurate tumor detection. Therefore, the dual-branch methods are proposed by fusing the global and local (lesion) features to enhance the model capability for tumor detection. Wang \emph{et al.} \cite{DLGNet} proposed DLGNet, a dual-branch lesion-aware neural network for colorectal lesion classification, which explicitly extracted the global and local features from the colorectal endoscopic images. Similarly, Basu \emph{et al.} \cite{RadFormer} proposed a Transformer-based network termed RadFormer to integrate global and local attention branches of the gallbladder lesion, outperforming competitive results in gallbladder tumor detection tasks. 
%Cai \emph{et al.} \cite{VANet} proposed VANet of Transformer architecture to focus on the polyp areas and enhance the perception of the polyp boundaries, leading to significant improvements in polyp segmentation performance. 
Although the existing methods have integrated the global and local (lesion) information and made notable improvements in cancer diagnosis, there are still some significant challenges that need to be addressed. One of the key challenges is their inability to accurately locate the lesion region in endoscopic images. This can be attributed to the fact that the lesion (foreground) region and the background often appear quite similar, making it difficult to differentiate them. Moreover, the existing dual-branch networks typically concatenate the global and local features directly, and then feed them into a module like a transformer to generate a fusion feature \cite{RadFormer}. These approaches are still challenging to fully capture the complementary features from the global and local branches, leading to inadequately utilizing the advantages of feature fusion mechanism which could further improve the laryngo-pharyngeal tumor detection performance.

To address the aforementioned challenges, we propose a Segment Anything Model-Guided Fusion Network (SAM-FNet), a novel dual-branch architecture to capture the global and local features for accurate laryngo-pharyngeal tumor detection. Specifically, it consists of five key components, a SAM-guided lesion location (SLL) module, a global feature extractor (GFE) module, a local feature extractor (LFE) module, a GAN-like feature optimization (GFO) module, and a classifier. Particularly, to accurately locate the lesion region, we leverage the SAM of powerful segmentation capabilities to segment tumors from laryngo-pharyngeal endoscopic images. To fully capture the discriminative information between the global and local branches, we propose a GAN-like optimization module to better learn the complementary feature representations. Extensive experiments on our collected two datasets from the First Affiliated Hospital of Sun Yat-sen University (FAHSYSU) and the Sixth Affiliated Hospital of Sun Yat-sen University (SAHSYSU) demonstrate that our proposed network SAM-FNet can achieve competitive results, surpassing the state-of-the-art counterparts on the LPC detection task. Overall, our main contributions can be summarized as follows: 
\begin{itemize}
\item[$\bullet$] We propose a novel SAM-guided fusion network for laryngo-pharyngeal tumor detection, which is the first to utilize a dual-branch network architecture for this specific task.

\item[$\bullet$] We introduce the SAM, with the advantage of powerful object segmentation capabilities, to segment the laryngo-pharyngeal endoscopic images for accurately locating the lesion region.

\item[$\bullet$] We utilize a GAN-like feature optimization module to further capture the complementary features between the global and local branches.  

\item[$\bullet$] Extensive experiments on the two datasets of FAHSYSU and SAHSYSU demonstrate that SAM-FNet can achieve competitive results, outperforming the state-of-the-art counterparts. 
\end{itemize}

\section{Methodology}
\label{sec:method}

\subsection{Overview}
\label{subsec:overview}
\begin{figure*}[tb]
    \centering
    \includegraphics[width=0.9\textwidth]{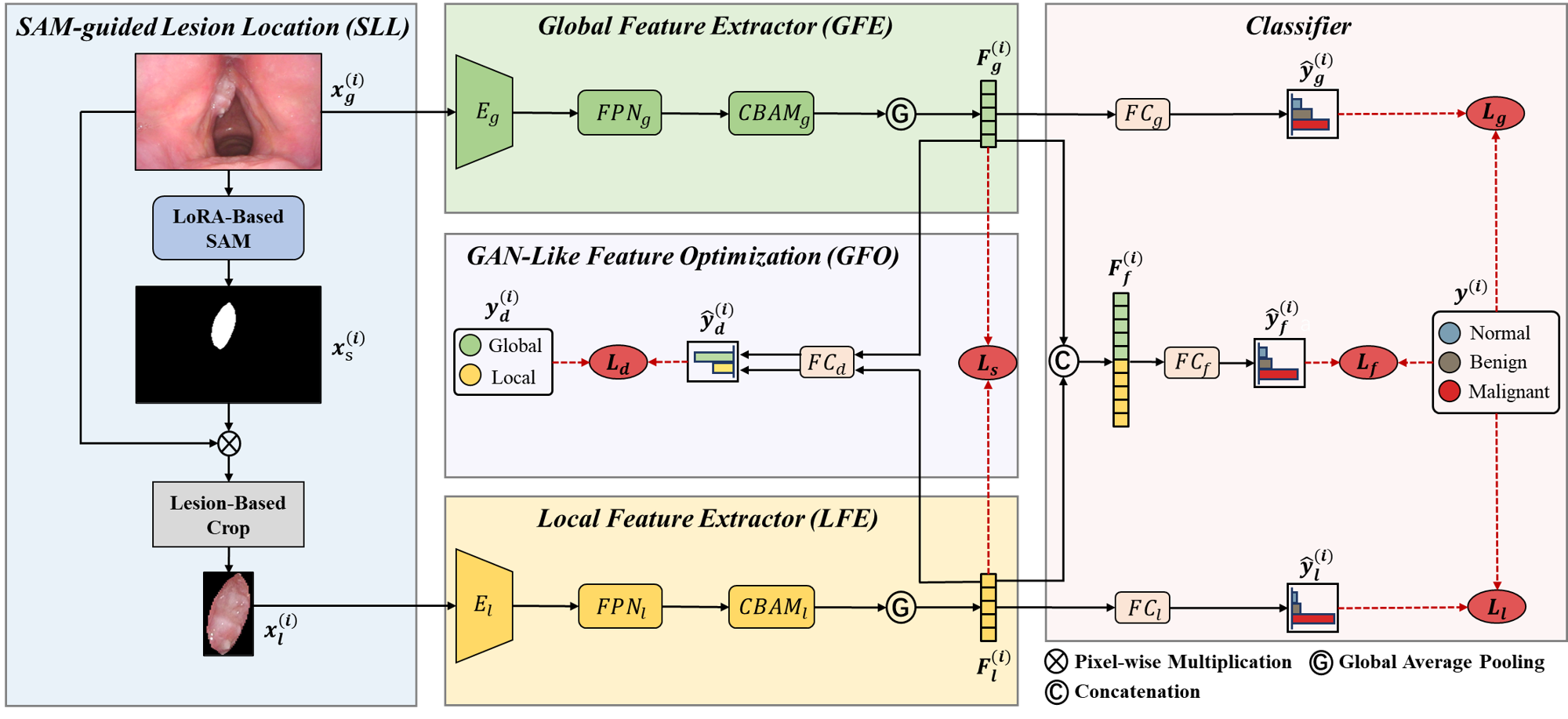}
    \caption{The architecture of the proposed SAM-FNet includes several key components: a SAM-guided lesion location (SLL) module to generate the lesion area image from the entire image; a global feature extractor (GFE) to extract features from the whole image; a local feature extractor (LFE) to derive features from the lesion region; a GAN-like feature optimization (GFO) module to align global and local features while differentiating their distributions; and a classifier that predicts based on global, local, and fused features.
    }
    \label{F.SAM-FNet}
\end{figure*}

Motivated by DLGNet \cite{DLGNet}, a dual-branch network can capture both global and local (lesion) information, enabling accurate detection of lesions. In this study, we propose a SAM-guided fusion network for laryngo-pharyngeal tumor detection, following dual-branch architecture \cite{DLGNet}, which is shown in Fig.~\ref{F.SAM-FNet}. Specifically, it consists of five parts: a SLL (see Section \ref{ssl}), a GFE (see Section \ref{global}), a LFE (see Section \ref{local}), a GFO (see Section \ref{gan}), and a classifier (see Section \ref{cls}). Unlike DLGNet, we introduce the SAM to accurately locate the lesion region. Additionally, we design a GAN-like feature optimisation module (GFO) to enhance global and local feature representations, thereby improving the model's ability to extract more discriminative features.

Formally, the dataset is denoted as $\mathcal{D} = \{ (x_g^{(i)}, y^{(i)}) \}_{i=1}^N$, where $N$ is the number of total laryngo-pharyngeal endoscopic images, $x_g^{(i)}$ represents a holistic laryngoscopic image, and $y^{(i)} \in \{0, 1, 2\}$ represents the labels for the three categories: normal laryngo-pharyngeal tissues (normal for short), benign tumors (benign for short), and malignant tumors (malignant for short). The lesion area image $x_l^{(i)}$ corresponding to $x_g^{(i)}$, is obtained through the SLL module and share the same label $y^{(i)}$ as $x_g^{(i)}$. Following the paradigm of DLGNet, we adopt a multi-task learning framework with several loss functions tailored for each branch, which can leverage domain-specific information from complementary tasks, enhancing prediction accuracy and generalization for each task \cite{komeda2017computer}. Specifically, global, local, and fused feature representations are fed into their respective classifiers for final prediction. The discrepancy between these predictions and the ground truth labels is measured by cross-entropy losses denoted as $\mathcal{L}_g$, $\mathcal{L}_l$, and $\mathcal{L}_f$, respectively. Additionally, a GAN-like loss $(\mathcal{L}_s + \mathcal{L}_d)$, comprising a similarity loss $\mathcal{L}_s$ (to align global and local features) and a binary cross-entropy loss $\mathcal{L}_d$ (to differentiate between global and local feature distributions), is introduced to further refine the feature representations, ensuring the capture of complementary and distinctive characteristics. In this study, we employ a joint learning scheme, where the total objective loss is defined as:
\begin{equation}
    \mathcal{L}_t = \mathcal{L}_f + \alpha \mathcal{L}_g + \beta \mathcal{L}_l + \gamma (\mathcal{L}_s + \mathcal{L}_d)
\end{equation}
where the weights $\alpha$, $\beta$, and $\gamma$ are trade-off hyperparameters of each component loss.

\subsection{SAM-guided Lesion Location}
\label{ssl}
To improve the model's capability in extracting lesion features, recent studies have concentrated on dual-branch networks \cite{DLGNet,RadFormer,VANet}. 
Specifically, these networks typically use segmentation methods, such as Mask R-CNN \cite{he2017mask}, to identify and crop the lesion region as the local input \cite{DLGNet}. However, these approaches often face challenges in accurately locating lesions because of the high similarity between the lesions and the surrounding tissues. To address this issue, we employ the SAM, which has demonstrated excellent performance in detecting pixel-level objects. However, the SAM is primarily trained on annotations for natural images, which differ significantly from medical images in domain characteristics. This domain discrepancy can lead to a substantial drop in performance when directly applied to medical images \cite{segmentexp, segmentmed}. Therefore, we utilize the low-rank-based (LoRA) fine-tuning strategy to make the SAM adapted to the laryngo-pharyngeal endoscopic images. Specifically, we freeze the SAM's image encoder and apply the LoRA layers to the query and value projection layers of the multi-head attention mechanism within each transformer block of the encoder, while keeping all parameters in the mask decoder and prompt encoder trainable. Furthermore, the rank of the LoRA layers is empirically set to 4 to balance efficiency and performance. 

After the fine-tuning process, we employ the LoRA-based SAM model to segment the entire image $x_g^{(i)}$, obtaining the corresponding lesion mask $x_s^{(i)}$, which can be defined to be
\begin{equation}
    x_s^{(i)} = \mathcal{F}_{SAM}(x_g^{(i)}),
\end{equation}
where $\mathcal{F}_{SAM}(\cdot)$ represents the function of the LoRA-based SAM that generates the lesion mask from the input image. Next, we apply pixel-wise multiplication between the entire image $x_g^{(i)}$ and the corresponding lesion mask $x_s^{(i)}$. This process isolates the lesions, effectively removing background information. The resulting image, containing only the lesions, is then used for lesion-based cropping to extract the lesion area image $x_l^{(i)}$. The entire process can be defined as follows:
\begin{equation}
    x_l^{(i)} = Crop(x_g^{(i)} \odot x_s^{(i)}),
\end{equation}
where $Crop(\cdot)$ denotes lesion-based cropping function, and $\odot$ indicates the pixel-wise multiplication. 

\subsection{Global Feature Extractor}
\label{global}
To capture comprehensive context information, we propose the GFE, designed to extract global features from the whole LPC image. The GFE includes three components of an encoder ($E_g$), a Feature Pyramid Network ($FPN_g$), and a Convolutional Block Attention Module ($CBAM_g$). Specifically, we utilize the ResNet-50 \cite{resnet} with the initial weights transferred from the model zoo trained on ImageNet as the backbone of the encoder $E_g$. Here, we remove the final fully connected layer. Then, we utilize the Feature Pyramid Network \cite{fpn} to further learn the fine-to-coarse granular information from feature maps output by $E_g$. The FPN architecture, as proposed in the work by Lin et al. \cite{fpn}, requires four-scale feature maps as inputs. In this study, we leverage the feature representations extracted from the last four blocks of the ResNet-50 model as the four-scale inputs to the FPN. As the FPN is initially proposed for object detection task, it derives five projection heads from the top-down pathway for further processing. However, this study is a classification task which merely requires one projection head for subsequent process. Thereby, we retains the bottom projection head which preserves the rich semantic information of images. To reduce the model's focus on irrelevant information among feature maps, we employ the CBAM \cite{cbam} with a channel attention module and a spatial attention module to highlight the inter-channel and inter-spatial relationships of features pertinent to laryngo-pharyngeal tumors. Finally, a global average pooling is utilized to map the feature maps into a global feature vector $F_g^{(i)}$, which can be defined to be
\begin{equation} \label{g_eq}
    F_g^{(i)} = \mathcal{F}_{GFE}(x_g^{(i)}),
\end{equation}
where $\mathcal{F}_{GFE}(\cdot)$ represents the global feature extractor.

\subsection{Local Feature Extractor}
\label{local}
As shown in the middle bottom part of Fig. \ref{F.SAM-FNet}, the Local Feature Extractor (LFE) employs the identical network architecture as the GFE described in Section \ref{global} to learn the local feature information, that is, the feature information of the lesion region. This allows us to effectively capture the relevant characteristics of the localized lesion area, which can be important for accurately identifying and analyzing the medical condition. Specifically, the LFE consists of  an encoder ($E_l$), a FPN ($FPN_l$), and a CBAM ($CBAM_l$). Importantly, the LFE does not share the parameters with the GFE, thereby enhancing its ability to learn the lesion-specific information. Formally, given a lesion area image $x_l^{(i)}$, passing it through the LFE processing, a local feature vector $F_l^{(i)}$ is obtained, which is defined to be
\begin{equation} \label{l_eq}
    F_l^{(i)} = \mathcal{F}_{LFE}(x_l^{(i)}),
\end{equation}
where $\mathcal{F}_{LFE}(\cdot)$ represents the local feature extractor.

\subsection{GAN-Like Feature Optimization}
\label{gan}
The GAN-Like Feature Optimization (GFO) module is proposed to improve the learning of complementary features between the global and local branches of the SAM-FNet architecture. Inspired by the Generative Adversarial Networks (GAN) framework \cite{gan}, the GFO module introduces an adversarial training process to better align the features learned by the global and local branches. Concretely, the GFO utilizes two loss functions of the similarity loss $\mathcal{L}_s$ and the discrimination loss $\mathcal{L}_d$, which is shown in the middle part of Fig. \ref{F.SAM-FNet}. The similarity loss aims to align the global feature vector $F_g^{(i)}$ (see Eq.~\ref{g_eq}) and the local feature vector $F_l^{(i)}$ (see Eq.~\ref{l_eq}), promoting the GFE and the LFE to better capture the coherent feature information. On the contrary, the discrimination loss encourages the model to learn the complementary feature information between the GFE and LFE. In this context, the similarity loss is calculated specifically for laryngo-pharyngeal endoscopic images that contain tumors. The labels for these images are either benign (with a label value of 1) or malignant (with a label value of 2). The similarity loss is calculated by cosine function, which is defined to be
\begin{equation}
    \mathcal{L}_s = \frac{1}{N_t} \sum_{i=1}^{N_t} \left( 1 - \frac{F_g^{(i)} \cdot F_l^{(i)}}{\| F_g^{(i)} \| \| F_l^{(i)} \|} \right),
\end{equation}
where $N_t$ is the number of laryngo-pharyngeal endoscopic images with tumor labels, and $\|F_g^{(i)}\|$ and $\|F_l^{(i)}\|$ are the Euclidean norms of the global and local feature vectors, respectively. However, simply reducing the distance between $F_g^{(i)}$ and $F_l^{(i)}$ in the feature space may lead to the loss of their specific advantages. Therefore, inspired by the adversarial approach in GAN, we introduce a discriminator network. The objective of this discriminator is to ensure that $F_g^{(i)}$ and $F_g^{(i)}$ maintain their unique characteristics: the global features should retain their broad contextual understanding, and the local features should focus on detailed information of tumors. Specifically, the discriminator, denoted as $FC_d$, is a fully connected layer that processes either the global feature vector $F_g^{(i)}$ or the local feature vector $F_l^{(i)}$. The output of the discriminator is given by:
\begin{equation}
    \hat{y}^{(i)}_{d} = \mathcal{F}^d_{FC}(F_g^{(i)} || F_l^{(i)}),
\end{equation}
where $||$ denotes "or". The binary cross-entropy loss $\mathcal{L}_d$ is then computed by comparing this predicted probability $\hat{y}^{(i)}_{d}$ with the ground truth label $y^{(i)}_d$, which is defined as:
\begin{equation}
    \mathcal{L}_d = \text{BinaryCrossEntropyLoss}(\hat{y}^{(i)}_{d}, y^{(i)}_d),
\end{equation}
where $y^{(i)}_d = 0$ indicates a global feature, and $y^{(i)}_d = 1$ indicates a local feature. This loss function optimizes the discriminator's ability to correctly classify the features as either global or local, ensuring that each feature retains its distinctive characteristics. With this adversarial optimization strategy, SAM-FNet can effectively extracts discriminative features between the global and local branches, enhancing the richness and complementarity of the fused feature representation.

\subsection{Classifier}
\label{cls}
In the training phase, following the DLGNet \cite{DLGNet}, we adopt the multi-task learning framework that incorporates three cross-entropy losses from the global, local, and fused branches. This framework has the potential to enhance prediction accuracy and generalization across each task \cite{komeda2017computer}. As shown in the right part of Fig.~\ref{F.SAM-FNet}, the fused feature vector $F_f^{(i)}$ is obtained by concatenating the global feature vector $F_g^{(i)}$ (see Eq. \eqref{g_eq}) and the local feature vector $F_l^{(i)}$ (see Eq. \eqref{l_eq}). The vectors $F_g^{(i)}$, $F_l^{(i)}$, and $F_f^{(i)}$ are then passed through their respective classifiers, denoted as fully connected layers $FC_g$, $FC_l$, and $FC_f$. This produces the probability distributions $\hat{y}_g^{(i)}$, $\hat{y}_l^{(i)}$, and $\hat{y}_f^{(i)}$, respectively, which can be defined as follows:
\begin{align}
    \hat{y}_g^{(i)} &= \mathcal{F}_{FC}^g(f_g^{(i)}), \\
    \hat{y}_l^{(i)} &= \mathcal{F}_{FC}^l(f_l^{(i)}), \\
    \hat{y}_f^{(i)} &= \mathcal{F}_{FC}^f(f_f^{(i)}),
\end{align}
% where $\mathcal{F}_{FC}^g(\cdot)$, $\mathcal{F}_{FC}^l(\cdot)$, and $\mathcal{F}_{FC}^f(\cdot)$ represent the function of fully connected layers of $FC_g$, $FC_l$, and $FC_f$, respectively. 
To optimize the SAM-FNet model, this study employs three separate cross-entropy loss functions, denoted as $\mathcal{L}_g$, $\mathcal{L}_l$, and $\mathcal{L}_f$, corresponding to the respective fully connected layers of $FC_g$, $FC_l$, and $FC_f$. The formulation of the loss functions can be expressed as follows:
\begin{align}
    \mathcal{L}_g &= \text{CrossEntropyLoss}(\hat{y}_g^{(i)}, y^{(i)}),\\
    \mathcal{L}_l &= \text{CrossEntropyLoss}(\hat{y}_l^{(i)}, y^{(i)}),\\
    \mathcal{L}_f &= \text{CrossEntropyLoss}(\hat{y}_f^{(i)}, y^{(i)}).
\end{align}

In the inference phase, inspired by ensemble learning, we generate the final output by averaging the values of $\hat{y}_g^{(i)}$, $\hat{y}_l^{(i)}$, and $\hat{y}_f^{(i)}$. This approach leverages the strengths of different branches, helping to mitigate individual prediction errors and enhance overall model robustness and accuracy.

\section{Experiment}
\label{sec:experiment}
\subsection{Experiment Settings}
\subsubsection{Datasets}

\begin{table}[!tb]
\centering
\caption{Statistic description of the FAHSYSU and SAHSYSU datasets.}
\label{dataset_distribution}
\begin{tabular}{@{}lcccc@{}}
\toprule
\multirow{2}{*}{} & \multicolumn{2}{c}{FAHSYSU (Total=25,256)} & \multicolumn{2}{c}{SAHSYSU (Total=2,788)} \\ \cmidrule(lr){2-3} \cmidrule(lr){4-5}
                  & NBI                 & WLI                 & NBI                & WLI                 \\ \midrule
Normal            & 695                 & 7,310                & 9                  & 2,202                \\
Benign            & 1,488                & 3,332                & 27                 & 218                 \\
Malignant         & 5,954                & 6,477                & 99                 & 233                 \\
Total             & 8,137                & 17,119               & 135                & 2,653                \\ \bottomrule
\end{tabular}
\end{table}

% This study was approved by the Ethics Committee of FAHSYSU (approval no. [2021]416). 
All LPC endoscopic images are provided by two hospitals, including the FAHSYSU and SAHSYSU. All the laryngo-pharyngeal endoscopic images were collected during routine clinical practice using standard laryngoscopes (ENF-VT2, ENF-VT3, or ENF-V3; Olympus Medical Systems, Tokyo, Japan) and imaging systems (VISERA ELITE OTV-S190, EVIS EXERA III CV-190, Olympus Medical Systems) at an original resolution of $512\times512$ pixels. Notably, our dataset includes laryngoscopic images captured in both narrow-banding imaging (NBI) mode and white light imaging (WLI) mode. Furthermore, all malignant tumors were pathologically confirmed. Tumors were annotated by experienced laryngologists, followed by cross-checking and expert review for quality control. The statistic description of laryngo-pharyngeal endoscopic images in each dataset is presented in Table~\ref{dataset_distribution}.
\begin{itemize}
    \item \textbf{FAHSYSU:} The FAHSYSU as the internal dataset contains 25,256 images, with 8,137 in NBI mode and 17,119 in WLI mode. It was used for model training, validation, and internal testing.
    \item \textbf{SAHSYSU:} The SAHSYSU as the external dataset contains 2,788 images, with 135 in NBI mode and 2,653 in WLI mode. It was only used for external testing without any data leakage on model training.
\end{itemize}

% All the endoscopic images are provided by two hospitals, including the First Affiliated Hospital of Sun Yat-sen University (FAHSYSU) and the Sixth Affiliated Hospital of Sun Yat-sen University (SAHSYSU). The specific number of laryngoscopic images in each dataset is presented in Table 1. The FAHSYSU dataset was used for model training, validation, and internal testing, while the SAHSYSU dataset was used for external testing. All the laryngoscopic images were collected during routine clinical practice using standard laryngoscopes (ENF-VT2, ENF-VT3, or ENF-V3; Olympus Medical Systems, Tokyo, Japan) and imaging systems (VISERA ELITE OTV-S190, EVIS EXERA III CV-190, Olympus Medical Systems) at an original resolution of 512x512 pixels. Notably, our dataset includes laryngoscopic images captured in narrow-band imaging (NBI) mode, which compensates for the limitations of white light imaging (WLI) by enhancing the clarity and recognizability of microvasculature. 

% The laryngoscopic images in the dataset were categorized based on clinical experience into normal, benign, and malignant, with all malignant tumors pathologically confirmed. Laryngoscopic images containing tumors were annotated by experienced laryngologists using labelme to delineate tumor boundaries, followed by cross-checking and expert review for quality control.

\subsubsection{Evaluation Metrics}
Accuracy, precision, recall, and $F_1$ score were utilized as evaluation metrics for the classification of laryngo-pharyngeal tumors. Moreover, we employed Dice coefficient (Dice) to analyze the segmentation performance of the LoRA-based SAM.

\subsubsection{Implementation Details} \label{imp_details}
All experiments were conducted on a computing server equipped with NVIDIA RTX A6000 GPU and the CUDA version used was 11.8. We writted all code in Python 3.10.14 and Pytorch 2.2.0 environment. The first stage involved fine-tuning based on the SAM, closely adhering to the settings described in SAMed \cite{samed}. However, due to differences in input data (CT images) and the number of prediction categories, several adjustments were made to adapt the model to our specific task: (1) We resized the input images to a size of \(224 \times 224\). (2) The model predicted a single category, distinguishing between foreground and background. (3) We empirically assigned a weight of 0.9 to the Dice loss and 0.1 to the cross-entropy loss.

In the second stage, the parameters of the LoRA-based SAM were frozen, and only the subsequent dual-branch network was trained. The masks generated by the LoRA-based SAM were used to crop the lesion region, serving as the input to the LFE. If no masks were detected (e.g., in the cases where the image did not contain tumors), a central region of a size of \(256 \times 256\) was cropped from the entire image. Both the holistic images and the lesion region images were resized to the same size. We then trained all networks using the Stochastic Gradient Descent (SGD) as the optimizer, with a learning rate of 0.003, a momentum of 0.9, a weight decay of 5e-4, and a mini-batch size of 256 for 60 epochs. We also utilized exponential learning rate decay with an exponent of 0.965. In order to improve the model's generalization and robustness, we have also employed the online data augmentation, such as random affine transformations, horizontal flipping and color jittering. Empirically, we set the hyperparameters $\alpha$, $\beta$, and $\gamma$ to 1.0, 0.3, and 0.01 in this study, respectively.  

\begin{table*}[!tb]
\centering
\caption{Experiment performance comparison with state-of-the-art counterparts on the FAHSYSU and SAHSYSU datasets.}
\label{model_performance_combined}
\begin{tabular}{llcccccccccc}
\toprule
\multirow{2}{*}{Dataset} & \multirow{2}{*}{Method} & \multicolumn{4}{c}{Overall results} & \multicolumn{3}{c}{Recall for different classes} \\ \cmidrule(lr){3-6} \cmidrule(lr){7-9}
& & Accuracy & Precision & Recall & $F_1$ score & Normal & Benign & Malignant \\
\midrule
\multirow{6}{*}{FAHSYSU} 
& ResNet \cite{resnet}            & {89.45\%} & {86.14\%} & 85.13\% & 85.52\% & {93.82\%} & 67.89\% & 93.68\% \\
& EfficientNet \cite{Efficientnet} & 89.31\% & 85.76\% & {85.71\%} & {85.69\%} & 93.16\% & 71.23\% & 92.75\% \\
& ViT \cite{ViT}                  & 87.74\% & 82.93\% & 84.57\% & 83.65\% & 90.96\% & {71.89\%} & 90.85\% \\
& RadFormer \cite{RadFormer}      & 87.01\% & 82.87\% & 82.42\% & 82.63\% & 89.08\% & 65.55\% & 92.61\% \\
& DLGNet \cite{DLGNet}            & 88.95\% & 85.04\% & 84.60\% & 84.80\% & 91.18\% & 68.47\% & {94.13\%} \\
& SAM-FNet (Ours)                        & \textbf{92.14\%} & \textbf{89.57\%} & \textbf{88.68\%} & \textbf{89.08\%} & \textbf{93.95\%} & \textbf{75.81\%} & \textbf{96.27\%} \\
\midrule
\multirow{6}{*}{SAHSYSU} 
& ResNet \cite{resnet} & 91.07\% & 80.18\% & 82.37\% & 81.22\% & 94.81\% & 67.05\% & 85.24\% \\
& EfficientNet \cite{Efficientnet} & 87.88\% & 74.69\% & 81.58\% & 77.50\% & 90.52\% & 68.97\% & 85.24\% \\
& ViT \cite{ViT} & 89.67\% & 78.14\% & 79.95\% & 78.68\% & 94.03\% & 67.82\% & 78.01\% \\
& RadFormer \cite{RadFormer} & 86.80\% & 71.57\% & 78.64\% & 74.61\% & 90.30\% & 63.98\% & 81.63\% \\
& DLGNet \cite{DLGNet} & 86.76\% & 72.89\% & 80.80\% & 75.96\% & 89.20\% & 67.05\% & 86.14\% \\
& SAM-FNet (Ours) & \textbf{92.29\%} & \textbf{82.71\%} & \textbf{84.52\%} & \textbf{83.59\%} & \textbf{95.58\%} & \textbf{69.73\%} & \textbf{88.25\%} \\
\bottomrule
\multicolumn{9}{l}{$^{\mathrm{1}}$The best performance is in \textbf{bold}.}
\end{tabular}
\end{table*}

\subsection{Experiment Results}
\subsubsection{Baselines}
To demonstrate the effectiveness of our proposed SAM-FNet, we compared our method with three types of state-of-the-art classification methods in both internal dataset and external dataset. Including CNN-based methods (ResNet and EfficientNet), Transformer-based method (ViT), and dual-branch methods (RadFormer and DLGNet).

\begin{itemize}
    \item \textbf{ResNet} \cite{resnet}: ResNet is a deep convolutional neural network that employs a residual connections to enable the training of very deep networks, effectively improving image recognition accuracy.
    \item \textbf{EfficientNet} \cite{Efficientnet}: EfficientNet is a convolutional neural network that scales depth, width, and resolution using a compound scaling method, optimizing accuracy and efficiency for image classification tasks.
    \item \textbf{ViT} \cite{ViT}: ViT is a vision transformer model that applies the transformer architecture, originally designed for natural language processing, to image classification by treating images as sequences of patches.
    \item \textbf{RadFormer} \cite{RadFormer}: RadFormer is a dual-branch network that uses a transformer architecture to combine global and local feature maps for accurate Gallbladder Cancer detection from Ultrasound images.
    \item \textbf{DLGNet} \cite{DLGNet}: DLGNet is also a dual-branch network that integrates contextual lesion information by learning global and local features for colon lesions classification.
\end{itemize}

\subsubsection{Internal Dataset Results}
Test data from the FAHSYSU dataset were initially randomly selected by laryngologists, with the data partitioned according to individual patients. For hyperparameter tuning, the remaining data were divided into training and validation sets in a 90\% to 10\% ratio. The detailed distribution is shown in Table~\ref{FAHSYSU_data_distribution}, with a total of 16,222 images in the training set, 1,806 images in the validation set, and 7,229 images in the test set. It should be noted that for a fair comparison, we downloaded the code for all baseline methods from their open-source repositories and retrained them on the FAHSYSU dataset using the same training settings as our proposed SAM-FNet.

\begin{table}[!tb]
    \centering
    \caption{Data distribution for training, validation, and test sets of the FAHSYSU dataset.}
    \label{FAHSYSU_data_distribution}
    \begin{tabular}{lccc}
    \toprule
           & Training & Validation & Test \\ \midrule
    Normal & 5,134 & 571 & 2,301 \\
    Benign & 3,276 & 366 & 1,178 \\
    Malignant & 7,812 & 869 & 3,750 \\
    Total & 16,222 & 1,806 & 7,229 \\ 
    \bottomrule
    \end{tabular}
\end{table}

Table~\ref{model_performance_combined} shows the evaluation results of our method compared with other state-of-art counterparts on the FAHSYSU dataset. It can be observed that the SAM-FNet achieves promising results, which can be up to 92.14\%, 89.57\%, 88.68\%, and 89.08\% in terms of accuracy, precision, recall, and $F_1$ score, respectively. Notably, in terms of single-class recall, SAM-FNet achieves 75.81\% for benign tumors and 96.27\% for malignant tumors.

It is noteworthy that our proposed SAM-FNet outperforms other state-of-the-art counterparts by a significant margin. Specifically, SAM-FNet surpasses the second-best counterpart by 2.69\% in accuracy and 2.97\% in recall. Additionally, SAM-FNet achieves a 3.92\% higher recall for benign tumors compared to the second-best counterpart.

\begin{figure}[!tb]
    \centering
    \includegraphics[width=0.45\textwidth]{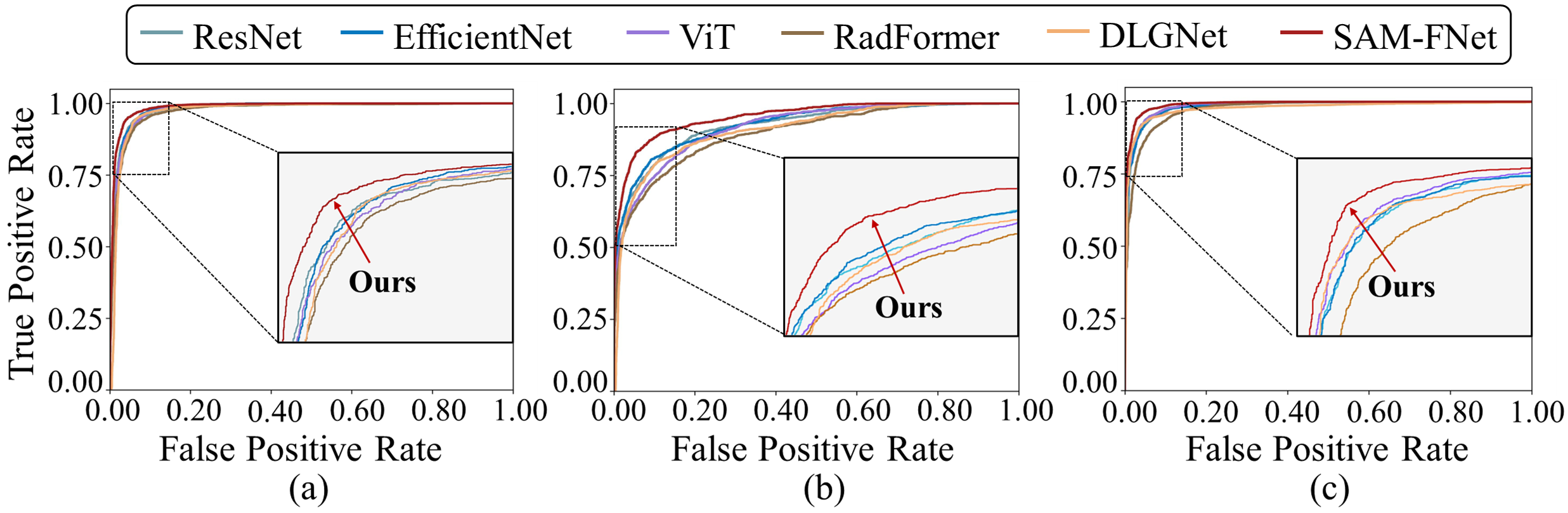}
    \caption{Receiver Operating Characteristic (ROC) curves for experiments results on the FAHSYSU dataset: (a) Normal, (b) Benign, (c) Malignant. Our proposed SAM-FNet, represented by red line, achieves the best classification performance across all classes.}
    \label{F.roc}
\end{figure}

Moreover, as shown in Fig.~\ref{F.roc}, we plotted the Receiver Operating Characteristic (ROC) curve for each category (normal, benign, malignant) compared with state-of-the-art counterparts. It can be observed that our method, represented by the red line, achieves the largest area under the curve (AUC) across all classes. This demonstrates superior classification performance of our proposed SAM-FNet. Notably, SAM-FNet significantly surpasses other counterparts in accurately classifying benign tumors. This may be due to the small size of benign tumors and their blurred borders with surrounding normal tissue, which make it difficult for other compared counterparts to extract lesion-related features. However, the SLL module in our method can accurately segment the contours of the lesion and crop the lesion region image, allowing the subsequent LFE to capture more discriminative tumor features.

\subsubsection{External Dataset Results}
The external dataset may suffer from inconsistent data distribution with the internal dataset, posing challenges for the model's accuracy in tumor detection. To assess the generalization performance of our method compared to state-of-the-art counterparts, we conducted comparative experiments on the SAHSYSU dataset. The detailed data distribution for this dataset is presented in Table~\ref{dataset_distribution}.

Table~\ref{model_performance_combined} also presents the performance of our proposed SAM-FNet in comparison with state-of-the-art counterparts on the external SAHSYSU dataset. The results clearly show that SAM-FNet achieves superior performance across all metrics. Specifically, SAM-FNet achieves an accuracy of 92.29\%, a precision of 82.71\%, a recall of 84.52\%, and an $F_1$ score of 83.59\%. In terms of sing-class recall, SAM-FNet reaches 69.73\% for benign tumors and 88.25\% for malignant tumors.

In comparison with other cutting-edge methods, SAM-FNet shows remarkable improvements in recall. Specifically, it surpasses the second-best counterpart by 2.15\% in overall recall. Even though DLGNet achieves a recall of 86.14\% for malignant tumors, it is still 2.11\% lower than our method.

\subsubsection{Ablation Experiments}
To verify the effectiveness of each component in the SAM-FNet, including GFE, LFE, and GFO module, we conducted ablation experiments on the FAHSYSU dataset.

\begin{table}[!tb]
\centering
\caption{The experiment results of ablation experiments.}
\label{ablation_study}
\scalebox{0.8}{
\begin{tabular}{cccccccc}
\toprule
% \multirow{1}{*}{Variants} & \multirow{1}{*}{GFE} & \multirow{1}{*}{LFE} & \multirow{1}{*}{GFO} \\ 
% \multicolumn{4}{c}{Overall results} & \multicolumn{3}{c}{Recall for different classes} \\ 
% \cmidrule(lr){5-8}
% \cmidrule(lr){9-11}
Variant & GFE & LFE & GFO & Accuracy & Precision & Recall & $F_1$ score \\
% & Normal & Benign & Malignant \\
\midrule
\textit{V1} & \checkmark & & & 89.45\% & 86.14\% & 85.13\% & 85.52\% \\
% & 93.82\% & 67.89\% & 93.68\% \\
\textit{V2} & & \checkmark & & 84.52\% & 79.38\% & 80.07\% & 79.64\% \\
% & 88.90\% & 62.39\% & 88.93\% \\
% \textit{V3} & \checkmark & \checkmark & & & & 90.62\% & 87.72\% & 86.44\% & 86.88\% & \textbf{96.01\%} & 69.06\% & 94.24\% \\
% \textit{V4} & \checkmark & \checkmark & \checkmark & & & 90.95\% & 88.20\% & 86.74\% & 87.33\% & {95.13\%} & {69.97\%} & 95.12\% \\
\textit{V3} & \checkmark & \checkmark & & {91.38\%} & {88.41\%} & {87.99\%} & {88.19\%} \\
% & {92.46}\% & \textbf{75.81\%} & {95.71\%} \\
\textit{V4 (SAM-FNet)} & \checkmark & \checkmark & \checkmark & \textbf{92.14\%} & \textbf{89.57\%} & \textbf{88.68\%} & \textbf{89.08\%} \\
% & \textbf{93.95\%} & \textbf{75.81\%} & \textbf{96.27\%} \\
\bottomrule
\multicolumn{8}{l}{$^{\mathrm{1}}$The best performance is in \textbf{bold}.}
\end{tabular}
}
\end{table}

Table~\ref{ablation_study} shows the ablation experiments results. It is observed that the variants \textit{V3} and \textit{V4}, which utilize both global and local feature extractors, outperform the variants \textit{V1} and \textit{V2}, which use only a single feature extractor. Specifically, variants \textit{V3} and \textit{V4} show improvements of at least 1.93\% in each evaluation metric. It means that the fusion of global and local features can effectively improve the laryngo-pharyngeal tumors detection performance. As for the variants \textit{V1} and \textit{V2}, the variant of \textit{V1} achieves significantly better performance than the variant \textit{V2} across all evaluation metrics, with improvements of at least 4.75\%. This demonstrates the importance of global semantic understanding of holistic laryngo-pharyngeal images for their classification. From the last row of the table, it can be observed that the introduction of GFO module helps the variant to achieve promising performance. The variant \textit{V4} improves by 0.78\%, 1.16\%, 0.69\%, and 0.89\% in terms of accuracy, precision, recall, and $F_1$ score, respectively, compared with the variant \textit{V3}. This results indicate that the variant with the GFO module outperforms other variants without it detecting tumors.

% In particular, the \textit{V4} achieves a recall of 96.27\% for malignant tumors, which is 0.56\% higher than that of the \textit{V3}. This indicates that the variant with the GFO module outperforms other variants without it in detecting malignant tumors.

% Compared with \textit{V3}, the model of \textit{V4} is with an additional FPN module, and its recall for benign and malignant tumors improves about 0.9\%. This suggests that the fusion of multi-scale information can help to facilitate tumors detection performance. As for the attention mechanism, it can enhance the model's capture of lesion-related features while suppressing the interference of irrelevant information, which potentially improve the classification performance. For example, compared with the model of \textit{V4}, the model of \textit{V5}, which includes an additional attention module, has a higher overall recall value of 87.99\%, improving by 1.25\%. Furthermore, the recall value for benign tumors, rises from 69.97\% to 75.81\%, achieving a significant improvement. 

\subsection{Visualization Analysis}

\begin{figure}[!tb]
    \centering
    \includegraphics[width=0.45\textwidth]{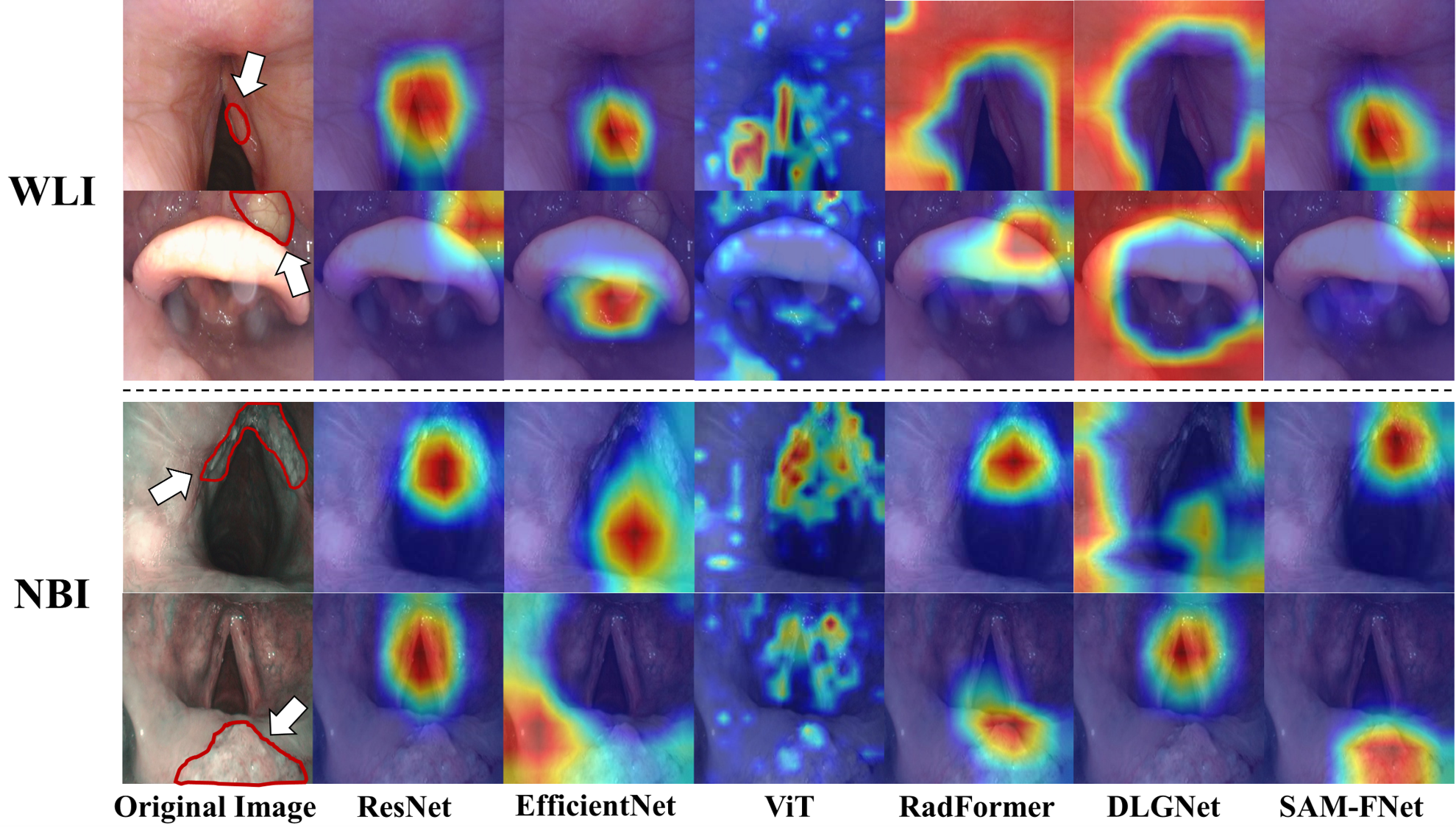}
    \caption{Illustrations of the Grad-CAM visualization for tumor images in both NBI and WLI modalities. Compared with other state-of-the-art counterparts, SAM-FNet is able to focus on and highlight effective tumor characteristics more precisely.}
    \label{F.cam}
\end{figure}

\begin{figure}[!tb]
    \centering
    \includegraphics[width=0.45\textwidth]{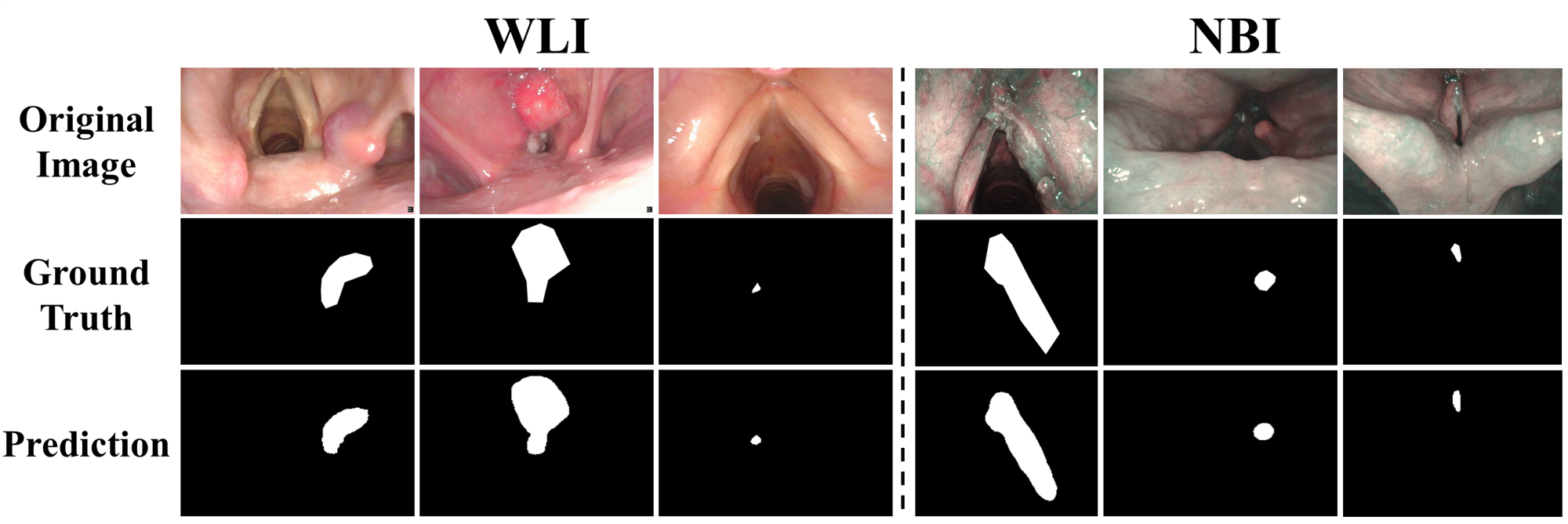}
    \caption{Illustrations of predicted lesion masks generated by the LoRA-based SAM within the SLL module in both NBI and WLI modalities. The predicted masks produced by our LoRA-based SAM demonstrate a high level of correspondence with the ground truth masks across these imaging modalities.}
    \label{F.sam_vis}
\end{figure}

In order to demonstrate the effectiveness of our proposed architecture intuitively, Gradient-weighted Class Activation Mapping (Grad-CAM) \cite{gradcam} was applied to show which parts the model pay attention to. Fig.~\ref{F.cam} shows examples of Grad-CAM visualizations on laryngoscopic images. It is observed that SAM-FNet correctly identifies tumors, and is able to focus on and highlight effective tumor characteristics more precisely than other state-of-the-art counterparts.

Furthermore, we conducted experiments and visualizations using the LoRA-based SAM on the FAHSYSU dataset to validate its effectiveness in lesion localization. Specifically, the LoRA-based SAM achieved a mean Dice coefficient of 0.5918 for benign tumors and 0.7966 for malignant tumors. In particular, the LoRA-based SAM demonstrates promising results in the segmentation of malignant tumors, potentially enhancing the lesion feature extraction capabilities of the LFE. To visually demonstrate the effectiveness of the LoRA-based SAM, we also present several segmentation results in Fig.~\ref{F.sam_vis}. The results indicates that the LoRA-based SAM exhibits strong segmentation performance on tumors, even when tumors are very small.

\section{Conclusion}
\label{sec:conclusion}
In this study, we propose a novel SAM-guided fusion network (SAM-FNet), a dual-branch architecture specifically designed for laryngo-pharyngeal tumor detection. The SAM-FNet consists of five key components: a SAM-guided Lesion Location (\emph{i.e.}, SLL) , a Global Feature Extractor (\emph{i.e.}, GFE), a Local Feature Extractor (\emph{i.e.}, LFE) module, a GAN-Like Feature Optimization (\emph{i.e.}, GFO) , and a classifier. To capture the critical lesion information in laryngo-pharyngeal endoscopic images, we introduce the SLL module that leverages powerful object segmentation capabilities to accurately identify and segment the lesion regions for subsequent feature extraction. This ensures that the network focuses on the most relevant areas of the tumor, improving the overall detection performance. Furthermore, to better capture the comprehensive characteristics of laryngo-pharyngeal tumors, we propose the GFO module, which utilizes a GAN-like mechanism to learn the complementary features between the global and local branches of the network. By fusing the global and local representations, the model can gain a more thorough understanding of the tumor's morphological and textural features, leading to enhanced classification accuracy. We evaluate the proposed SAM-FNet on two datasets, FAHSYSU as the internal dataset and SAHSYSU as the external dataset, and the results demonstrate its effectiveness. The SAM-FNet achieves an overall accuracy of 92.14 \% and 92.29 \% on the FAHSYSU and SAHSYSU datasets, respectively, surpassing state-of-the-art approaches and showcasing its competitive performance on the laryngo-pharyngeal tumor detection task.

\end{document}